# WTTFNet: A Weather-Time-Trajectory Fusion Network for Pedestrian Trajectory Prediction in Urban Complex

Ho Chun Wu, Esther Hoi Shan Lau, Paul Yuen, Kevin Hung, John Kwok Tai Chui and Andrew Kwok Fai Lui
School of Science and Technology, Hong Kong Metropolitan University, Kowloon, Hong Kong Special Administrative Region, China
E-mail: { ahcwu; hslau; chyuen; khung; jktchui; alui }@hkmu.edu.hk

*Abstract*— Pedestrian trajectory modelling in an urban complex is challenging because pedestrians can have many possible destinations, such as shops, escalators, and attractions. Moreover, weather and time-of-day may affect pedestrian behavior. In this paper, a new weather-time-trajectory fusion network (WTTFNet) is proposed to improve the performance of baseline deep neural network architecture. By incorporating weather and time-of-day information as an embedding structure, a novel WTTFNet based on gate multimodal unit is used to fuse the multimodal information and deep representation of trajectories. A joint loss function based on focal loss is used to co-optimize both the deep trajectory features and final classifier, which helps to improve the accuracy in predicting the intended destination of pedestrians and hence the trajectories under possible scenarios of class imbalances. Experimental results using the Osaka Asia and Pacific Trade Center (ATC) dataset shows improved performance of the proposed approach over state-of-the-art algorithms by 23.67% increase in classification accuracy, 9.16% and 7.07% reduction of average and final displacement error. The proposed approach may serve as an attractive approach for improving existing baseline trajectory prediction models when they are applied to scenarios with influences of weather-time conditions. It can be employed in numerous applications such as pedestrian facility engineering, public space development and technology-driven retail.

## I. Introduction

Predicting pedestrian trajectories in crowd scenario is essential in smart city. It has numerous applications such as self-driving cars [1], smart road crossings and intelligent retail [2]. KB models describe pedestrian dynamics using physical, social or psychological rules. Pioneer KB models are Social Force model [3] and collision avoidance [4]. Deep learning (DL) approaches leverage extensive observations. They can be mainly categorized into Recurrent Neural Network (RNN) [5] [6], Convolutional Neural Network (CNN) [7], Transformer (TF) [8], Generative Adversarial Network (GAN) [9] [10] [11]. Most recent research focuses on Social-awareness incorporated deep neural network architectures [10] and graph convolutional network (GCN) [12] to further improve performance.

While much attention is directed towards modelling the trajectory in outdoor scenarios with applications to autonomous vehicles, this paper focuses on modelling the pedestrian trajectory within an urban complex. Recently, Indoor Pedestrian Trajectory Generator (IPTG) [13] was reported, which uses a GAN based approach to generate trajectories for a fictional conference scenario. Han et al. [14] employed trajectory clustering in modeling pedestrian flow for indoor design space. D'Orazio et al. [15] simulated the pedestrian flow of a building using agent-based model with proximity and exposure time based rules to estimate the spread of Coronavirus Disease (COVID-19) in building. However, there is few existing literature about indoor pedestrian trajectory modelling under the influence of weather and time-of-day, a.k.a. weather-time (WT) condition. Xue et al. [16] studied the modelling of pedestrian movement in a train station and proposed a Pedestrian Trajectory Prediction method by LSTM with Automatic Route Class Clustering (PoPPL). It employed *k*-mean clustering to label pedestrian trajectories followed with subsequent LSTM based intent classification and trajectory prediction. However, the train station dataset only contained video lasting for 30 minutes with same weather and it mainly serves the purpose of transportation.

Weather-time (WT) conditions refer to weather and time-of-day variations. An objective of this paper is to study the effect of weather and time-of-day for pedestrian movement pattern in urban complex. Typical indoor environment, such as residential apartments, offices, factories, etc., are single functional premises. Individuals usually share common location-of-interest (LOI), i.e. going home/going to work. In contrast, pedestrian behavior in urban complexes exhibits much more randomness as the pedestrians could have different destinations to functional objects [17] that serves a wide range of purposes, such as retail, shopping malls, office accommodations, and business functions. Previous studies [18], [19] suggested that weather has an impact in affecting pedestrian behavior. In particular, bad weather may discourage consumers from shopping. Also, adverse weather conditions may lead to delays or cancellations of public transportation services [20], which affects pedestrian traffic. Time-of-day will affect commuter traffic and hence pedestrian flow [21], [22]. This study aims to improve understanding on how the weather and time-of-day influence the choice of destination and hence the trajectories of pedestrians, which will help to facilitate flow management [23] and intelligent retail [2]. With the increasing popularity of multimodal transportation in large metropolises to decrease reliance on private cars and greenhouse emission, many urban complexes are designed with multimodal transportation [24] capabilities. They serve as interconnection points to facilitate seamless transfers between buses and trains. Examples are Osaka station (Osaka, Japan) [24] and Chatswood interchange shopping mall (Sydney, Australia).

Three practical issues may arise in modelling the pedestrian trajectory under different weather-time (WT) conditions in urban complex are i) appropriate preprocessing and feature selection, ii) effective fusion, iii) choice of clusters under the effect of different WT conditions.

First, the format of weather information may not directly fit for use and require appropriate preprocessing and feature selection. Directly concatenating this information to the deep neural network may even confuse the classifier and lead to inferior performance. For instance, Time-of-day information is commonly available as numeric values and the classifier may perceive it as ordinal, i.e. 9 o'clock is larger than 8 o'clock, which is not logical at all.

Second, it is not trivial on where and how to fuse the WT information. For example, direct concatenation of one-hot encoded WT information to the raw pedestrian trajectories does not yield satisfactory performance.

Third, although the use of trajectory prediction guided by pedestrian intent have been reported before, it is mainly used to predict the pedestrian's intent for road crossing in outdoor scenarios [25] involving pedestrian-vehicle interaction. Unlike the road crossing scenario, where pedestrians will need to cross the road under different weather conditions, the pedestrian behavior in urban complex can be affected by weather, especially in destinations for retail and entertainment.

To overcome these challenges in improving the pedestrian trajectory prediction accuracies of baseline deep learning models, we propose a new weather-time-trajectory network for destination adapted pedestrian trajectory prediction (WTTFNet) that incorporates the effect of WT conditions for prediction of pedestrian trajectories, which is made up of four stages:

1. Weather-time (WT) Embedding: To tackle the issue of preprocessing and feature selection of WT information, a word embedding is used to encode the WT information and it has the advantage to be further optimized according to the final loss function.
2. Novel WTTF architecture: Motivated by the rationale that the weather-time conditions may influence the decision, the proposed WTTF architecture fuse the preliminary pedestrian intent probabilities with WT embedding. Inspired by the good performance of GMU [26] in fusing numerical and textual features, the GMU is adopted for the bimodal fusion.
3. Deep supervision [27] is used to co-train the preliminary and final classifiers together using auxiliary and final loss functions. While the preliminary pedestrian intent probabilities provide supervisory signals to train the baseline classifier, the final loss function optimizes the whole architecture. The Focal Loss [28] is used to cater for possible class imbalance.
4. Finally, a new statistical test based on the Pearson's chi-squared $\chi^2$ statistic is designed to determine the minimum sample size required for each cluster. $p$-value can be computed to test the significance of the WT condition and determine whether to incorporate the WT information.

To illustrate the effectiveness of the proposed approach in improving a baseline pedestrian trajectory model, the public dataset obtained from Asia and Pacific Trade Center (ATC) [29] in Osaka is considered. It is an urban complex serving as a multimodal transportation hub, which connects the intercity ferry pier and Osaka metro line, as well as accommodating a trade center and multi-entertainment complex. Pedestrian trajectories obtained on a sunny (22nd May, 2013) and cloudy day (29th September, 2013) were used. There were roughly 1.5 times more pedestrians during peak hours in compared to off-peak hours. A significant log $p$-value of -104.8395 $\ll \log(0.05)$[1] is attained using the proposed statistical test, which suggests that there is significant deviation in pedestrian flow across weather and off/peak hours.

Experimental results show that the proposed WTTFNet surpasses state-of-the-art algorithm by reduction of 9.16 % and 7.07% in average displacement error (ADE) and final displacement error (FDE), respectively. It also improves the classification accuracy (ACC) and Cohen's Kappa (κ) of the baseline model (i.e. PoPPL) by 23.67% and 28.13%, respectively.

To study the role of weather and time-of-day in improving prediction performance, ablation test is performed to compare between the proposed WTTFNet with/without incorporation of weather-time information. Significant McNemar's test [30] $p$-value of $p$=0.0196<0.05 was attained, which suggests the improvement in classification accuracy from 71.5% to 71.95% after adding weather-time information was significant because of the large sample size of 28536 pedestrians.

Further analysis of the 3008 significant pedestrians identified by McNemar's test shows that an overall 5.47% (7.8m to 7.4m) and 7.58% (14.11m to 13.04m) improvement in ADE and FDE reduction were obtained for the significant 3008 pedestrians, and significant one-sided Mann–Whitney U test [32] $p$-values were attained for ADE ($p$=0.0203<0.05) and FDE ($p$=0.00533<0.05), respectively. This shows that weather-time information helps to improve prediction performance significantly for the 3008 cases considered. Overall, the ratio 3008 out of 28536 pedestrians was also statistically significant according to the McNemar's test, suggesting that these 3008 pedestrians showing significantly improved performance out of 28536 cases were very unlikely a random event. This suggests the proposed approach may serve as an attractive approach for incorporating WT information to improve pedestrian trajectory prediction and it also serves as a systematic approach to test the significance of WT conditions.

---
[1] A significance level of 0.05 is sought [31].

Finally, with the increasing popularity of multimodal transportation in large metropolises to decrease reliance on private cars and reduce greenhouse emission, understanding pedestrians' behavior in urban complex is increasingly important. Walking networks with interconnecting urban complexes will be increasingly prevalent to facilitate smooth transfers between different modes of transportation and contribute to the economic development of nearby areas. There are also numerous applications in public space development [33], evacuation planning [34], and advancements in technology-driven retail [2].

The rest of this paper is organized as follows. Section II presents a review on the background and related works, whereas the proposed WTTFNet is presented in Section III. In Section IV, experimental results and comparisons with state-of-the-art algorithms are presented. The proposed statistical test is also used to test the significance of weather-time effects. Finally, conclusion is drawn in Section V.

## II. BACKGROUND AND RELATED WORK

Pedestrian trajectory prediction (PTP) methods can be categorized according to input modality, network architecture, features, and prediction tasks [35] [36]. Traditionally, PTP is achieved using knowledge based methods such as social force [3] collision avoidance [4], kinetic models [37]. In the last decade, deep learning approaches have gained much popularity for its powerfulness in leveraging extensive observations. They can be mainly categorized into

1. Recurrent neural network (RNN): Examples are Long Short Term Memory (LSTM) [5], Social LSTM [38], Gated Recurrent Unit (GRU) and Conv-LSTM [39]. LSTM are renowned for its capability to handle sequence-to-sequence prediction. Social LSTM further extends LSTM to model social interactions. Conv-LSTM replaces the fully connected layers in conventional LSTM with convolutional layers, which enables the capturing of both spatial and temporal information for intent and trajectory prediction in [39].
2. Convolutional neural networks (CNN): The CNNs are usually used for PTP approaches that uses images/videos to predict the trajectories. CNN is used to extract spatial-temporal features [7] or skeleton keypoints (Piccoli et al. 2020) [40] for classifying pedestrian behaviour.
3. Transformer: VOSTN [8] used a variational one-shot transformer for trajectory prediction together with a cross-attention module to model the inter-relationship between trajectory and ego-motion. AgentFormer [10] integrated a transformer architecture with agent-aware attention mechanism and a conditional variational autoencoders (CVAE) based trajectory prediction framework.
4. Generative adversarial network (GAN): POI-GAN [41] used generative model that integrates interest point model, field of view angle, and observed trajectories, to produce projected pedestrian trajectories for future time frames. Social GAN [9] employs a LSTM model to capture temporal structure of individual pedestrian and a social pooling mechanism to aggregate pedestrian interactions. The resultant deep features are used to train the GAN.

Over the past 5 years, most research focuses on incorporation of Social-awareness [9] [10], or contextual information [25],[39] to improve prediction performance. Social-awareness approaches such as social LSTM (Alahi et al. 2016) [32], social GAN [9], Sophie [42], AgentFormer [10] etc., primarily center around predicting trajectories and modeling interactions among a fixed number of pedestrians based on social pooling mechanisms. GCN based approaches, such as Social Spatial Temporal Graph CNN (SSTGCNN) [43], which models pedestrian interactions as graphs and extract spatial-temporal feature from the graphs using convolutional operations.

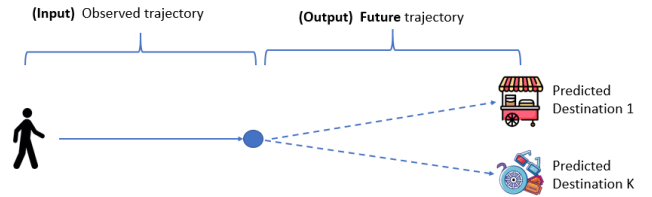

Fig. 1. A pedestrian trajectory prediction problem. The observed trajectory is used to predict the future trajectory and final destination in this paper.

Context-based approaches incorporates context information to predict pedestrian intent and use it to guide subsequent trajectory prediction [23],[33]. Typical pedestrian intent includes crossing road and other walking gestures [44]. These intents are predicted from video or LIDAR sequences. Examples of contextual information are road topology, maps, pedestrian attributes, road boundaries and ego-vehicle information [23], [33]. While much attention is directed towards modelling the trajectory in outdoor scenarios with applications to autonomous vehicles, this paper focuses on modelling the pedestrian trajectory within an urban complex. In the next section, the proposed methodology will be discussed.

## III. METHODOLOGY

Fig. 1 shows an illustration of the pedestrian trajectory prediction problem, where the proposed WTTFNet predicts the final destination and future trajectory from partially observed trajectory, e.g. half of the trajectory in this paper. The proposed WTTFNet is made up of the following components:

1. Destination-driven clustering: It is used to label the pedestrian trajectories of the training set with destinations assigned by k-mean clustering for subsequent training of the intended-destination (ID) classifier.
2. The proposed statistical test based on the Pearson's chi-squared $\chi^2$ statistic is designed to determine the minimum sample size required for each cluster and determine whether to incorporate the WT information.

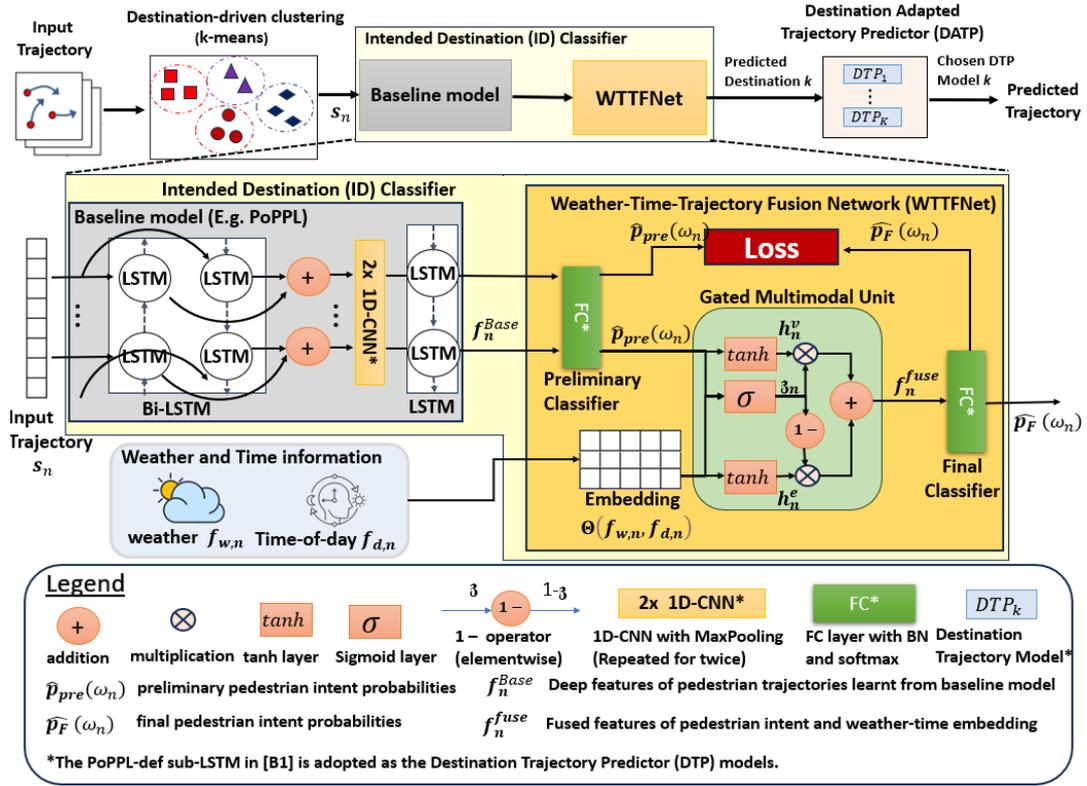

Fig. 2. The proposed WTTFNet. Tanh stands for Hyperbolic Tangent Function.

3. Intended-destination (ID) classifier: It predicts the final destination that occurs in future from an observed "historical" trajectory of the pedestrian. The training set is provided by the destination-driven clustering.
4. Destination adapted trajectory predictor (DATP): Each trajectory model is trained using the clustered trajectories, which targets only to the nearby area of the chosen destination. As an illustration, the PoPPL-def sub-LSTM [16] is adopted as the trajectory model. In general, other trajectory prediction models can be used.

### A. Destination-driven clustering module

In a pedestrian trajectory prediction problem, an observed trajectory $s_n$ for the $n-th$ pedestrian of length $L$ is used to predict the future $L'$ observations trajectory $\hat{z}_n$:

$$s_n = \{(x_{n,1}, y_{n,1}), \ldots, (x_{n,L}, y_{n,L})\} \quad (1)$$

$$\hat{z}_n = \{(\hat{x}_{n,L+1}, \hat{y}_{n,L+1}), \ldots, (\hat{x}_{n,L+L'}, \hat{y}_{n,L+L'})\}. \quad (2)$$

However, there are multiple possible destinations of a pedestrian and hence a destination-driven clustering will be beneficial for training destination-specific trajectory models. In the destination-driven clustering module, the end-point of all trajectories, i.e. $\Omega_{end}:\{(x_{n,L+L'}, y_{n,L+L'}), n = 1,2,\ldots,N\}$ from Eqn. (1b) are passed to the $k$-means algorithm to form clusters. The membership of an endpoint $(x, y)$ is sought by minimizing its distance from the centroids $\sum_{k=1}^{K} \sum_{(x,y) \in S_k} \|(x, y) - (\mu_{x,k}, \mu_{y,k})\|_2^2$, where $\Omega_k$ is the $k-th$ cluster and its centroid is updated as $\mu_k = [\mu_{x,k}, \mu_{y,k}]^T = \frac{1}{|\Omega_k|} \sum_{(x,y) \in \Omega_k} (x, y)$. $|\Omega_k|$ is the number of elements in $\Omega_k$. After assignment, each trajectory is labelled with the corresponding class from $\omega = 1, \ldots, K$ for sub-sequent training of the pedestrian intent classifier. The raw trajectories are cleaned and resampled so that the total duration of each trajectory is normalized to $T_o$.

The proposed approach also employs a statistical test to test the significance of each cluster and establish the minimum number of samples for each cluster (See Eqn. (12)). If a cluster is found to have insufficient number of samples, it can be merged to one of the clusters using an agglomerative clustering similarity measure, such as centroid linkage criterion $min. \| (\mu_{x,k}, \mu_{y,k}) - (\mu_{x,k_i}, \mu_{y,k_i}) \|_2^2$, where $(\mu_{x,k}, \mu_{y,k})$ is the centroid of the cluster to be merged and $(\mu_{x,k_i}, \mu_{y,k_i})$ are the remaining clusters. It is noted that other similarity measures can be employed. After the clusters have been computed, the training dataset for the ID classifier can be obtained as

Trajectory: $\quad s_n = \{(x_{n,1}, y_{n,1}), \ldots, (x_{n,L}, y_{n,L})\}, \quad (3a)$

Destination: $\quad \omega_n = 1, \ldots, K, \quad (3b)$

where $K$ is the total number of destinations.

TABLE I
STRUCTURAL DETAILS OF PROPOSED WTTFNET[1]

| Operation | Dimension (Input, Output) |
|---|---|
| (a) Preliminary Classifier (Eqn. (5b)) | |
| Fully connected (FC) + Batch Normalization +Softmax | ([2]$K_{LSTM}$,K) |
| (b) Weather-Time (WT) Embedding (Eqn. (4)) | |
| Embedding | ([3]C,128) |
| (c) Gated Multimodal Unit (Eqns. (6a) to (6d)) | |
| Tanh for Preliminary Classifier (Eqn. 6a) | (K,128) |
| Tanh for Embedding (Eqn. 6b) | (128,128) |
| Gate Neuron (Eqn. 6c) | (256,128) |
| 1 – operator, Multiplication ⊗, and Addition ⊕ | (128,128) |
| (d) Final Classifier (Eqn. (7)) | |
| FC + Batch Normalization+ Softmax | (128, K) |

[1]Batch size dimension is omitted in this table.
[2]$K_{LSTM}$ is the output dimension of the LSTM from PoPPL. Interested readers can refer to [16] for details of PoPPL.
[3]$C$ is the total number of weather-time conditions.

## IV. NOVEL WEATHER-TIME-TRAJECTORY NETWORK FOR DESTINATION CLASSIFICATION

Fig. 2 shows the proposed intended destination (ID) classifier, which comprises the weather-time (WT) embedding, baseline model (e.g. PoPPL) and the novel WTTFNet. First, a baseline model is used to learn the micro-level representation of the trajectory. Afterwards, a fully connected (FC) layer is used to learn a preliminary classifier of the destinations. The output preliminary ID class probabilities are then passed to the GMU for fusing with the WT embedding. The fused multimodal representation is passed to a final FC layer for training the final classifier. Both the preliminary and final classifier are co-optimized using the focal loss function. Here, the PoPPL is employed as the baseline model. In general, other trajectory models can be used.

More precisely, suppose there are $C_w$ weather conditions and $C_d$ different time-of-day and the total number of weather-time conditions are $C = C_w + C_d$. For example, in this paper, $C_w = 2$ (sunny/rainy) and $C_d = 2$ (off-peak/peak hours) are chosen. The proposed Weather-Time (WT) Embedding for the $n-th$ pedestrian is given as

$$e_{WT,n} = \Theta(f_{w,n}, f_{d,n}), \quad (4)$$

where Θ() is the embedding layer. $f_{w,n}$ and $f_{d,n}$ are the one-hot encodings describing the weather-time condition for the $n-th$ pedestrian. The preliminary ID class probabilities $\hat{p}_{pre}(\omega_n)$ can be obtained as the softmax probabilities from the preliminary classifier in Fig. 2. Batch normalization and softmax are performed after the FC layer. The preliminary intent probabilities $\hat{p}_{pre}(\omega_n)$ and preliminary classifier $f_n^C$ are given as

$$\hat{p}_{pre}(\omega_n) = \sigma_{Soft}(f_n^C), \quad (5a)$$

$$f_n^C = \phi_{BN}\left(FC\left(f_n^{base}\right)\right), \quad (5b)$$

respectively, where

$$\sigma_{Soft}(u) = \frac{1}{\sum_{k=1}^K e^{u_k}}[e^{u_1}, e^{u_2}, \dots, e^{u_K}]^T, \text{ and} \quad (5c)$$

$$\phi(u_k) = \frac{u_k - E(u_k)}{\sqrt{var(u_k)+\epsilon}} \times w_{\gamma,k} + w_{b,k} \quad (5d)$$

represent softmax operation and Batch normalization (BN), respectively. $f_n^C$ and $f_n^{base}$ are the output of the preliminary classifier and base model, respectively for the $n-th$ pedestrian. $\phi_{BN}(u) = [\phi(u_1), \phi(u_2), \dots, \phi(u_K)]^T$ is the batch normalization function. $FC(u) = W \cdot u$ is a fully connected layer with weights $W$ and $\sigma_{Soft}(u)$ is the softmax function. $w_{\gamma,k}$ and $w_{b,k}$ are learnable parameters for BN.

The preliminary pedestrian intent probabilities $\hat{p}_{pre}(\omega_n)$ and the WT embedding $\Theta(f_{w,n}, f_{d,n})$ are then fused at the GMU. The GMU is used to find an intermediate representation that fuses the two modalities (Arevalo et al. 2017), i.e. preliminary ID probabilities and WT embedding. First, the pedestrian intent probabilities and WT embedding are passed to individual tanh layers, each of which contains a neuron with hyperbolic tangent activation to encode the individual modalities. At the same time, a tied gate neuron learns the contribution of the two modalities, as shown in Fig. 2. The contributions $\mathfrak{z}_n$ and $(1 - \mathfrak{z}_n)$ obtained from the gate neuron will be multiplied in an elementwise manner to the output of the *tanh* layers of $\hat{p}_{pre}(\omega_n)$ and $\Theta(f_{w,n}, f_{d,n})$, respectively. A special feature of this gate unit is that $\mathfrak{z}_n$ supports multivariate weighting. To use this feature, the output dimension of the two *tanh* layers can be modified to a common dimension matching each other. Finally, the fused multimodal representation will be passed to the final classifier for predicting the final class probability.

More precisely, the GMU can be described using the following set of equations:

$$h_n^v = tanh\left(W_v \cdot \hat{p}_{pre}(\omega_n)\right), \quad (6a)$$

$$h_n^e = tanh\left(W_e \cdot \Theta(f_{w,n}, f_{d,n})\right), \quad (6b)$$

$$\mathfrak{z}_n = \sigma_{sgm}(W_\mathfrak{z} \cdot \left[\hat{p}_{pre}(\omega_n)^T, \Theta(f_{w,n}, f_{d,n})^T\right]^T), \quad (6c)$$

$$f_n^{fuse} = \mathfrak{z}_n \odot h_n^v + (1 - \mathfrak{z}_n) \odot h_n^e, \quad (6d)$$

where $h_n^v$ is the output of *tanh* layer for $\hat{p}_{pre}(\omega_n)$ for the $n^{th}$ pedestrian. $h_n^e$ is the output of *tanh* layer for Embedding. $\mathfrak{z}_n$ is the output of the gate neuron. $f_n^{fuse}$ denotes the fused representation. $\hat{p}_F(\omega_n)$ is the predicted pedestrian intent probabilities obtained from the final classifier and it is given as

$$\hat{p}_F(\omega_n) = \sigma_{Soft}\left(\phi_{BN}\left(FC_{M,K}\left(f_n^{fuse}\right)\right)\right). \quad (7)$$

Here, $tanh(u) = [tanh(u_1), tanh(u_2), \dots]^T$ and $tanh(u) = \frac{e^u - e^{-u}}{e^u + e^{-u}}$. $\sigma_{sgm}(u) = [\sigma_{sgm}(u_1), \sigma_{sgm}(u_2), \dots]^T$ and $\sigma_{sgm}(u) = \frac{1}{1+e^{-u}}$. The Hadamard product operator is denoted as ⊙. The set of unknown neural network weights to be learnt in the GMU are $\{W_v, W_e, W_\mathfrak{z}\}$, which corresponds to the weights of *tanh* layer for the preliminary pedestrian intent probabilities, *tanh* layer for the WT embedding and Gate Neuron, respectively. A common dimension $M$ is chosen for the two *tanh* layers in (6a) and (6b) so that they match the dimension of $\mathfrak{z}_n$. Finally, a FC layer $FC_{M,K}()$ with input dimension $M$ and output dimension $K$ is used to learn the final ID class probabilities. $\phi_{BN}(.)$ and $\sigma_{Soft}(.)$ are the batch normalization and softmax operations defined in (5c) and (5d), respectively. The preliminary and final classifiers will be jointly optimized as

$$L_T = (1 - \lambda_P)L_{focal}(\omega, \hat{\boldsymbol{p}}_F(\omega)) + \lambda_P L_{focal}(\omega, \hat{\boldsymbol{p}}_{pre}(\omega)). \quad (8)$$

Where for simplicity, we drop the subscript $n$ in (8). $L_{focal}(\omega, \hat{\boldsymbol{p}}_F(\omega))$ and $L_{focal}(\omega, \hat{\boldsymbol{p}}_{pre}(\omega))$ the losses for the final and preliminary classifiers, respectively. $\lambda_P$ is a parameter controlling the ratio of the two losses. It is chosen as $\lambda_P = 0.5$ in this paper. To cater for possible class imbalance, the focal loss (Lin et al.2017) is used

$$L_{focal}(\omega, \hat{p}) = -\frac{1}{NK}\sum_{n=1}^{N}\sum_{k=1}^{K} I_{k,n}\beta_k(1-\hat{p}_{k,n})^\gamma \times \log(\hat{p}_{k,n}), \quad (9)$$

where $\hat{p}_{k,n}$ is the predicted class probability for the $k-th$ class. $I_{k,n}$ is an indicator variable and $I_{k,n} = 1$ when the actual class is $K$. $\gamma$ is a focusing factor and $\beta_k \in [0,1]$ is a weighting factor. The final predicted class (i.e. intended destination) can be obtained as

$$\hat{\omega}_n = \max(\hat{p}_F(\omega_{n,1}), \hat{p}_F(\omega_{n,2}), \dots, \hat{p}_F(\omega_{n,K})), \quad (10)$$

where $\hat{p}_F(\omega_{n,k})$ is the softmax probability of the $k-th$ destination. $\hat{\boldsymbol{p}}_F(\omega) = [\hat{p}_F(\omega_{n,1}), \hat{p}_F(\omega_{n,2}), \dots, \hat{p}_F(\omega_{n,K})]^T$.

### B. Destination-adapted trajectory predictor module

After obtaining the final probabilities, the predicted trajectory can be obtained as

$$\hat{z}_n = DTP_{k=\hat{\omega}_n}(s_n)$$

where $\hat{z}_n$ is the predicted trajectory for the chosen class. $DTP_{k=\hat{\omega}_n}(s_n)$ is the chosen destination trajectory baseline model based on predicted class $\hat{\omega}_n$. The baseline model is chosen as sub-LSTM (PoPPL-def) in PoPPL [16] for the sake of comparison. The sub-LSTM (PoPPL-def) is an encoder-decoder LSTM with two hidden layers. In the next sub-section, the proposed statistical test will be presented.

### C. Statistical test of significance of weather-time conditions

The proposed statistical test can be used to establish the minimum required samples for each cluster and to quantify whether it is necessary to treat the pedestrian movement pattern in different periods and weathers as different groups and use different trajectory models to describe their behavior. More precisely, Table II shows a $K \times C$ contingency table summarizing the number of pedestrians arriving to $K$ destinations under $C$ different weather-time (WT) conditions. The following null hypothesis is proposed:

$$H_0: \text{The WT condition does not affect the choice of destination.} \quad (11)$$

If the null hypothesis is true, then the observed number of pedestrians should not deviate significantly from the expected counts across different WT conditions. According to the $\chi^2$ test, the minimum number of expected samples/trajectories required for each cluster $k$ under condition $c$ is

$$e_{kc} = \frac{l_K \times \underline{n}_C}{\underline{n}} \geq 5 \quad (12)$$

where $l_k = \sum_{c=1}^{C} l_{kc}$, $\underline{n}_c = \sum_{k=1}^{K} l_{kc}$ and $\underline{n} = \sum_{k=1}^{K}\sum_{c=1}^{C} l_{kc}$. $l_{kc}$ is the number of observed trajectories/samples in the $k-th$ destination and $c-th$ condition.

TABLE II
PROPOSED STATISTICAL TEST OF SIGNIFICANCE OF WEATHER-TIME CONDITIONS[1]

| Class\Condition | 1 | 2 | … | $C$ | Total |
|---|---|---|---|---|---|
| Class 1 | $l_{11}$ | $l_{12}$ | … | $l_{1C}$ | $l_1$ |
| ⋮ | ⋮ | ⋮ | ⋮ | ⋮ | ⋮ |
| Class $K$ | $l_{K1}$ | $l_{K2}$ | … | $l_{KC}$ | $l_K$ |
| Total | $\underline{n}_1$ | $\underline{n}_2$ | … | $\underline{n}_C$ | $\underline{n}$ |

[1] $K \times C$ Contingency table of number of pedestrian arrivals from different destinations $\omega = 1, \dots, K$ under different WT conditions $c = 1,2,\dots,C$.
[2] Example: Condition 1 can be chosen as Peak period + Sunny. Condition 2 can be chosen as off-peak+ sunny. Condition 3 can be chosen as peak + cloudy. Condition 4 can be chosen as off-peak + cloudy. Total number of conditions are $C = 4$.

Once the clusters are established, the test statistic for WT condition reads

$$\chi^2_{obs} = \sum_{c=1}^{C}\sum_{k=1}^{K} \frac{(o_{kc} - e_{kc})^2}{e_{kc}}, \quad (13)$$

where $o_{kc}$ is the actual observed number of pedestrians in condition $c$ and destination $k$. The $p$-value of the test is given as

$$p = Pr(\chi^2 \geq \chi^2_{obs,j}|H_0), \quad (14)$$

where the test statistic follows a $\chi^2$ distribution with $(C-1)(K-1)$ degree of freedom. At a significance level of 0.05 (Ross, 2020), the null hypothesis will be rejected when the $p$-value is smaller than 0.05 and it will suggest the difference between the proportion of pedestrian arrival under different conditions and origins is statistically significant.

## V. RESULTS AND DISCUSSIONS

For illustrative purposes, the Osaka Asia and Pacific Trade Center (ATC) dataset (Dražen et al. 2013) is considered. The Osaka ATC is a transportation hub linking the Sunflower intercity Ferry pier to the Osaka City Metro. It contains a multi-entertainment complex and a conference center. The trajectories were collected at 1/F of ATC using 3D range sensors. The full dimension is over 140 $m \times 60\ m$. Trajectories from 0900 to 2000 on 22nd May, 2013 (sunny) and 29th September, 2013 (cloudy) are chosen. Trajectories that are too short are removed (i.e. same cluster for origin and destination) as they may be a result of occlusion or tracking loss of the 3D range sensor. After resampling, the trajectory length $L + L' = 40$. The total number of pedestrian trajectories after pre-processing are 7329 on 22nd May, 2013 (sunny) and 21207 on 29th September, 2013 (cloudy) and respectively. Hence, the total number of pedestrian/ trajectories are 28536. Each

TABLE III
LIST OF INITIALIZATION CENTROIDS FOR OSAKA ATC CENTRE 1/F

|  | $K = 10$ | [1]$K = 9$ |
|---|---|---|
| Exit to ferry pier | Centroid 1 | Centroid 1 |
| Exit to ticket office | Centroid 3 | Centroid 3 |
| Information Kiosk | Centroid 2 | Centroid 2 |
| Central Square | Centroid 6 | Centroid 6 |
| Stairs to G/F | Centroid 10 |  |
| Stairs to 2/F and floor guide | Centroid 7 | Centroid 7 |
| Stairs to the mall | Centroid 8 | Centroid 8 |
| Exit to metro station | Centroid 9 | Centroid 9 |
| Stairs to the mall | Centroid 4 | Centroid 4 |
| Escalator | Centroid 5 | Centroid 5 |

[1]According to the statistical test in Table IV, classes 6 and 10 are merged.

TABLE IV
NUMBER OF PEDESTRIAN ARRIVALS DURING PEAK HOUR (12:00-16:59). OFF-PEAK, SUNNY AND CLOUDY FOR OSAKA ATC DATASET (K=10)

|  | A | B | C | D | Total |
|---|---|---|---|---|---|
| Class 1 | 645 | 601 | 1135 | 1722 | 4103 |
| Class 2 | 71 | 102 | 113 | 256 | 542 |
| Class 3 | 25 | 46 | 123 | 230 | 424 |
| Class 4 | 625 | 953 | 2010 | 3912 | 7500 |
| Class 5 | 75 | 106 | 281 | 445 | 907 |
| Class 7 | 126 | 186 | 439 | 667 | 1418 |
| Class 8 | 653 | 1044 | 1226 | 2303 | 5226 |
| Class 9 | 938 | 1072 | 2637 | 3436 | 8083 |
| Class 6 | 1 | 2 | 6 | 21 | **30** |
| Class 10 | 20 | 38 | 55 | 190 | 303 |
| Total | **3179** | 4150 | 8025 | 13182 | 28536 |
| Min. sample before merging ($K=10$) | Applying Eqn. (12) $e_{61} = \frac{l_6 n_1}{n} = \frac{(30)(3179)}{28536} = 3.34 < 5$ **(not satisfied)** | | | | |
| Min. sample after merging ($K=9$) | Applying Eqn. (12) $e_{61} = \frac{l_6 n_1}{n} = \frac{(30+303)(3179)}{28536} = 37.09 > 5$ **(satisfied)** | | | | |
| $\chi^2$ after merging ($K=9$) | 588.64 (degree of freedom 24) | | | | |
| log($p$-value) after merging ($K=9$) | $-104.8395 <$ log (0.05) (significant) | | | | |

A: cloudy and off-peak   B: cloudy and peak
C: sunny and off-peak    D: sunny and peak

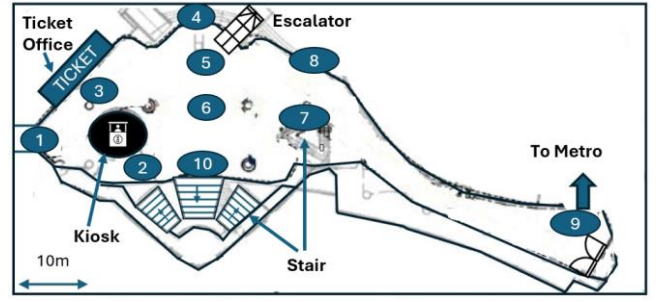

Fig. 3. Initialization centroids for $k$-means clustering, $K = 10$ classes added to 1/F floor plan of Osaka ATC Centre [29]. Important functional objects (i.e. ticket office, escalator, kiosk and stairs) are redrawn. The historical weather on 22nd May 2013 (sunny) and 29th September (cloudy), 2013 was obtained from [45].

pedestrian contains only 1 trajectory.

*A. Choice of clusters*

A general rule of thumb to choose the number of classes is to study the number of possible entrances/exits of the floor plan [16], [17]. Fig. 3 shows the floor plan of the Osaka ATC Center (1/F). Following this notion, key entrances and exits are chosen as the initialization centroids as in Fig. 3. Table 3 shows the list of initialization centroids.

*B. Statistical Analysis of time-of-day and weather conditions*

In this section, we shall test the significance of time-of-day and weather conditions using the proposed statistical test. Table 4 shows the number of observed pedestrian arrival during peak hour (12:00-16:59), off-peak, sunny and rainy conditions for $K = 10$. Using Eqn. (12), it was found that class $\omega_6$ does not meet the minimum sample requirement. Hence, using the centroid linkage criterion, $\omega_6$ is merged with $\omega_{10}$. The observed $\chi^2_{obs}$ computed using (13) is 588.64 (degree of freedom 24) and the log( $p$-value) is -104.8395, which is statistically significant under the typical significance level of 0.05 (Ross, 2020). This suggests there is a significant deviation in the pedestrian counts across the different clusters under the different conditions. Hence, the proposed approach should be used to model the pedestrian trajectory patterns under the different conditions. Next, we shall evaluate the performance of the various algorithms.

*C. Baseline and Metrics*

To evaluate the performance of the proposed approach, we compare the proposed WTTFNet with the following algorithms:

1. Linear Model: A simple linear model with a hidden layer (nn.linear in Pytorch) [46] is used to predict the trajectories.
2. Vanilla LSTM: The sub-LSTM in PoPPL-def is used. It employs an encoder-decoder LSTM with 2 hidden layers fitting all the trajectories. The implementation follows the Github codes [16].
3. PoPPL [16]: The sub-LSTM model is employed together with route class clustering. The implementation follows the Github codes. Following the previous statistical analysis, $K = 9$ destinations were chosen. Route class clustering divides all trajectories according to all combinations of all 9 origins and 9 destinations for training trajectory models.
4. Proposed WTTFNet: For fair comparison, we adopt the same baseline model as in PoPPL, as shown in Fig. 1. However, the proposed destination-driven clustering and proposed WTTFNet are used. Hyperparameters same as the authors are adopted for the PoPPL baseline model. For the number of destinations, $K = 9$ is chosen as in the previous analysis.

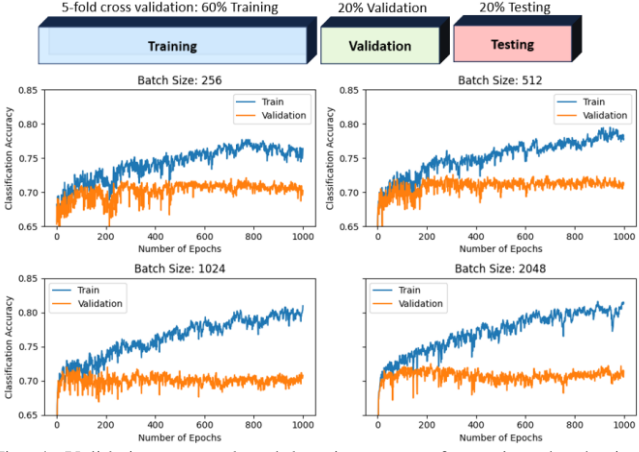

Fig. 4. Validation protocol and learning curves for various batch sizes. Stratified 5-fold cross validation (CV) is used.

For evaluating the quality of trajectory prediction, the average displacement error (ADE) is the average Euclidean distance between all the actual and all predicted coordinates over all trajectories. The FDE is the average Euclidean distance between the final destination of the predicted and actual trajectories. They are given as

$$ADE = \frac{1}{N_T L'}\sum_{n=1}^{N_T}\sum_{t=1}^{L'}\left\|\begin{pmatrix}x_{n,L+t}\\y_{n,L+t}\end{pmatrix}-\begin{pmatrix}\hat{x}_{n,L+t}\\\hat{y}_{n,L+t}\end{pmatrix}\right\|_2, \quad (15a)$$

$$FDE = \frac{1}{N_T}\sum_{n=1}^{N_T}\left\|\begin{pmatrix}x_{n,L+L'}\\y_{n,L+L'}\end{pmatrix}-\begin{pmatrix}\hat{x}_{n,L+L'}\\\hat{y}_{n,L+L'}\end{pmatrix}\right\|_2, \quad (15b)$$

where $\|.\|_2$ denotes the Euclidean distance. $N_T$ is the total number of testing samples. $(x_{n,t}, y_{n,t})$ is the actual coordinate and $(\hat{x}_{n,t}, \hat{y}_{n,t})$ is the predicted coordinate of the $n-th$ pedestrian's trajectory. The accuracy of the destination classification is evaluated using classification accuracy (ACC) and Cohen's Kappa ($\kappa$). They are given as

$$ACC = \frac{1}{N_{Test}}\sum_{k=1}^{K} CM[i,i], \quad (16a)$$

$$\kappa = \frac{N_T \sum_{i=1}^{K} CM[i,i] - \sum_{i=1}^{K} C_T[i]C_P[i]}{N_T^2 - \sum_{i=1}^{K} C_T[i]C_P[i]}, \quad (16b)$$

where $CM[i,j] = \sum_{n=1}^{N_{Test}} I(\omega_n = i \,\&\, \widehat{\omega_n} = j)$ is the total number of counts of having the actual class $i$ and predicted class $j$. $I(.)$ is the indicator function. $C_T[i] = \sum_{j=1}^{K} CM[i,j]$ and $C_P[j] = \sum_{i=1}^{K} CM[i,j]$. While classification accuracy is commonly used to describe the generic performance, Cohen's Kappa is used more frequently for scenarios with possible class imbalance. The following relative metrics, $rd$ are used to compare between different algorithms,

$$rd = \frac{(d - d_{REF})(-1)^m}{d_{REF} + \epsilon} \times 100\%, \quad (17)$$

where $d$ can be any metrics, such as the ADE, FDE, ACC and $\kappa$. $d_{REF}$ is the performance of the reference method. $\epsilon = 10^{-8}$ is a small constant added to denominator to avoid division by zero. $m$ is a parameter defining metric type. $m = 0$ is used for maximizing metrics with larger value indicating better performance, whereas $m = 1$ is used for loss metrics with smaller value indicating better performance.

### D. Experimental Setup

The Google Colab Tesla T4 Graphics Processing Unit (GPU) notebook with 16GB GPU memory and 17 GB of system memory is used for evaluation. In the experiment, each observed trajectory has a duration of 20 time-instants and an algorithm will predict the trajectory for the next 20 time-instants. Fig. 3 shows the validation protocol following the validation strategy in [16]. Stratified 5-fold cross validation (CV) is employed. Due to stratification and possible chances that the total number of samples is indivisible by 5, the number of samples across folds may vary slightly. Three-folds (~60%), one-fold (~20%), and one-fold (~20%) are used for training, validation, and testing, respectively.

TABLE V
TRAJECTORY PREDICTION PERFORMANCE OF VARIOUS ALGORITHMS.

| Metric/Model | Linear Model | Vanilla LSTM | [3]PoPPL (Original) | [3]Proposed WTTFNet | |
|---|---|---|---|---|---|
| | | | | (A) | (B) |
| ACC (%) | N/A | N/A | 58.18 | **71.50** | 71.95 |
| ADE(m) | 13.28 | 6.263 | 6.488 | **5.93** | 5.894 |
| FDE(m) | 22.84 | 10.687 | 11.266 | **10.42** | 10.315 |
| rACC (%) | N/A | N/A | Reference | 22.89% | 23.67% |
| rADE (%) | -104.69% | 3.47% | | 8.58% | 9.16% |
| rFDE (%) | -103.46% | 4.80% | | 6.13% | 8.12% |

A: Proposed WTTFNet without weather-time (WT) information
B: Proposed WTTFNet with WT information

[1] The relative improvement metrics are defined as in Eqn. (17).
[2] The Linear model and Vanilla LSTM does not contain a destination/route classifier. Hence, classification performance is not applicable.
[3] Both the PoPPL and the proposed WTTFNet select the specialized trajectory models using a classifier.

TABLE VI
TRAJECTORY PREDICTION PERFORMANCE OF VARIOUS ALGORITHMS

| Metric/Model | [3]PoPPL (Original) | [4]PoPPL+ Focal Loss | Proposed WTTFNet[5] | |
|---|---|---|---|---|
| | | | Without WT info | With WT info |
| ACC[1] | 58.18% | 68.84% | 71.50% | **71.95%** |
| $\kappa$[1] | 51.73% | 62.49% | 65.89% | **66.28%** |
| rACC[2] | Reference | 18.32% | 22.89% | **23.67%** |
| $r\kappa$[2] | | 20.80% | 27.37% | **28.13%** |

[1] ACC: Accuracy; $\kappa$: Cohen's Kappa
[2] Relative improvements of ACC and $\kappa$.
[3] PoPPL (Original): Original PoPPL optimized with loss function using cross entropy.
[4] PoPPL (FL): PoPPL optimized using focal loss (FL)
[5] Proposed WTTFNet: It is made up of PoPPL + FL + Deep supervision + WT information incorporated using GMU
[6] Since performance may vary across different baseline models, the PoPPL is adopted as the baseline model of the proposed approach for fair comparison. In general, other baseline models can also be applied.

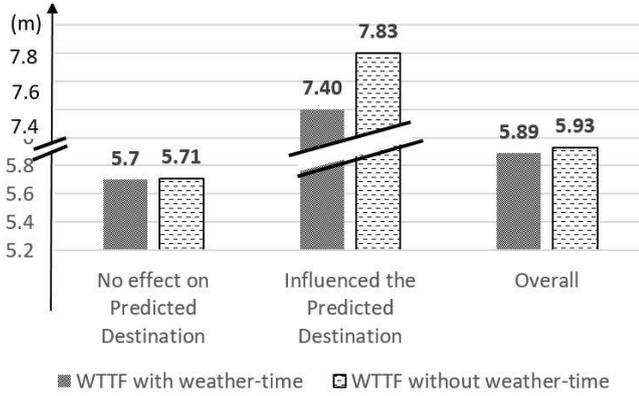

Fig. 5. Average Displacement Error (ADE) of the proposed approach with/without the incorporation of weather-time information. Significant reduction in ADE can be observed (7.8m to 7.4m) for the significant 3008 pedestrians out of all 28536 pedestrians.

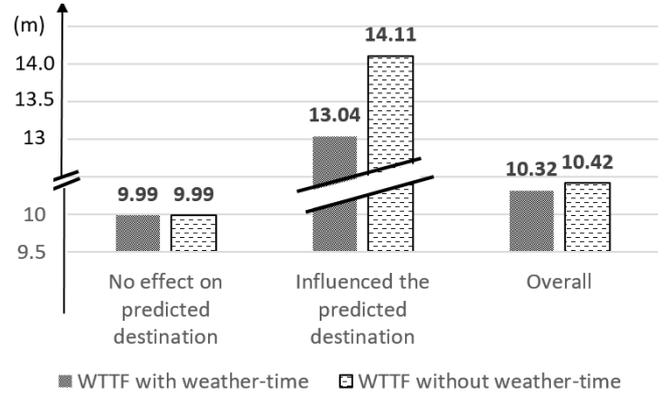

Fig. 6. Final Displacement Error (FDE) of the proposed approach with/without the incorporation of weather-time information. Significant reduction in FDE can be observed (14.11m to 13.04m) for the significant 3008 pedestrians out of all 28536 pedestrians.

### D.1 Batch Size Selection and Stopping Criterion

Fig. 4 shows the training and validation curves for the proposed WTTFNet under batch sizes 256, 512, 1024 and 2048. For batch size 256, the validation curve is quite noisy and fluctuates rapidly and hence it is not considered. For batch sizes 512, 1024 and 2048, the training accuracy starts to level off around epoch 100 but the validation accuracy remains roughly around a certain range. This suggests more epochs do not necessarily lead to better validation performance. Hence, 1000 epochs are chosen as stopping criterion. Overall, batch size 1024 attained the lowest variance in validation accuracy and hence it is chosen. For each CV fold, the model obtained at the epoch attaining the best validation accuracy is chosen and is used to evaluate the testing data.

### D.2 Hyperparameters

Hyperparameters same as the PoPPL are adopted for the baseline classifier and trajectory models. Dropout parameter of 0.5 and hidden size of 128 are adopted. For the proposed WTTFNet, the weighing factor in the focal loss is chosen as $\boldsymbol{\beta} = [\beta_1, \beta_2, ..., \beta_K]^T$, $\beta_k = (\frac{N/N_k}{\sum_{k=1}^{K} N/N_k})$, where $N$ is the total number of training samples, $N_k$ is the number of training samples of class $k$, $K$ is the total number of classes. The focusing parameter is chosen as $\gamma = 2$. The ratio between the preliminary and final loss in (8) is chosen as $\lambda_P = 0.5$.

### E. Experimental results

In this section, the proposed WTTFNet is compared against various algorithms. Since the proposed WTTFNet can be attached to arbitrary deep neural network baseline models, the PoPPL is adopted as baseline for illustration. In general, other deep neural network based intent classifier, such as transformers, can be adopted as the baseline model. Since the PoPPL is a technique that combined clustering and LSTM, we also compared with Vanilla LSTM.

Table 5 shows the overall performance of all algorithms. The proposed WTTFNet performed better than the original PoPPL, Vanilla LSTM and the linear model for all cases considered. Particularly, the proposed WTTFNet surpasses the original PoPPL 23.67% in classification accuracy, 9.16% reduction in ADE and 7.07% reduction in FDE. Significant $p$-values of ($p < 10^{-16}$) are attained for improvement in classification accuracy (McNemar's test [30]), ADE and FDE (one-sided Mann–Whitney U tests [32]).

### E.1 Ablation Test

To study the incremental contribution of each component of the proposed novel WTTFNet and show the role of weather and time-of-day in improving the prediction, we consider quantitative and qualitative analyses, which are shown in Table 6 and Fig. 5, respectively. Table 6 shows a comparison of the proposed WTTFNet under two different settings, (with/without WT information). Overall, there are two major factors that lead to the improvement of WTTFNet over existing PoPPL:

1. WTTFNet without WT information (second last column of Tables 5 and 6): Even when the GMU is bypassed and WT information is not supplied, the deep supervision used in the WTTFNet is useful in refining the preliminary pedestrian intent probabilities using both the preliminary and final classifiers optimized using auxiliary and final loss functions based on focal loss, irrespective of whether WT information is incorporated. This leads to improved classification accuracy (Table 6), reduction in ADE and FDE (Table 5).

2. WTTFNet with WT information (final column of Tables 5 and 6): It can be seen that the best performance (highest classification accuracy, lowest ADE and FDE) can be attained after incorporation of WT information into the proposed WTTFNet.

Comparing between the two different settings (with/without WT information) of the proposed WTTFNet, it was found that the classification accuracy increased from 71.5% to 71.95% after adding WT information. Significant $p$-value (Ross, 2020) ($p = 0.0196 < 0.05$) was attained for the McNemar's test, suggesting the statistical significance in the improvement of overall classification accuracy. This suggests the 3008 pedestrians out of 28536 were found to have significant improvement after the WT information was incorporated, this prompted us to further analyze those two different groups of pedestrians. With these evidences, this motivates us to further analyze the role of weather and time-of-day in next subsection.

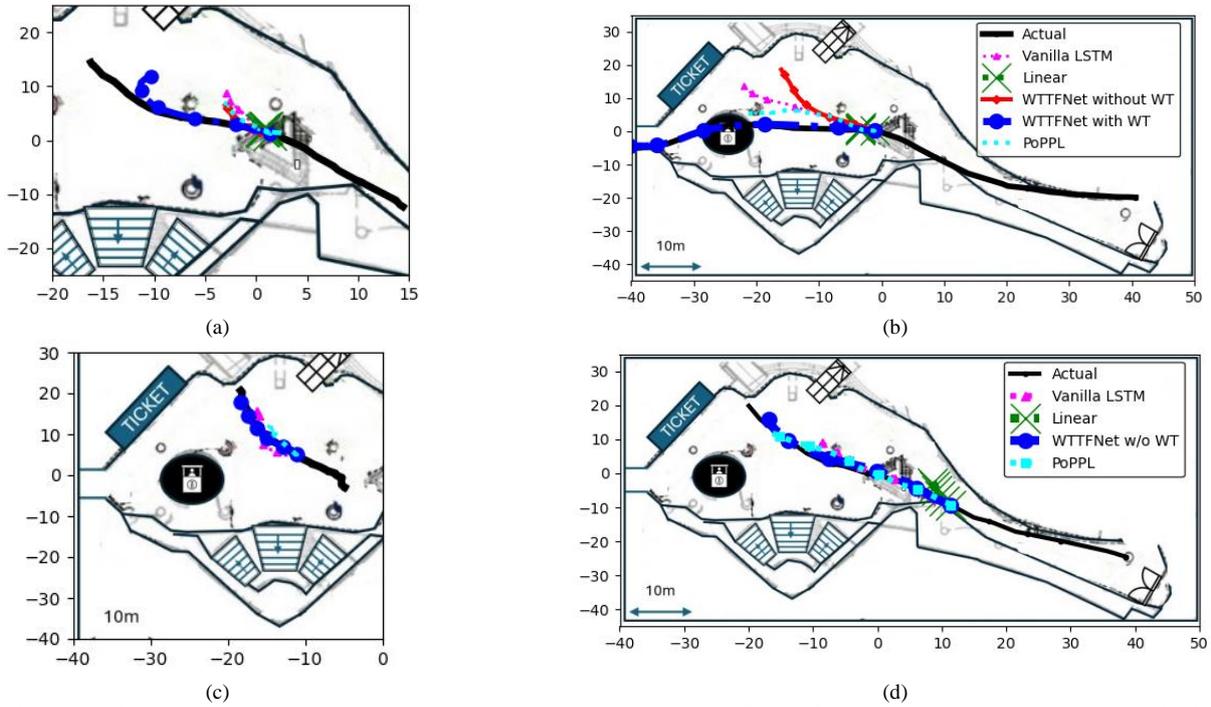

Fig. 7. Illustration of predicted trajectories, where the weather-time condition has (a,b) significant influence on destination (chosen from the 3008 significant pedestrians), and (c,d) no influence on destination (chosen from remaining pedestrians). The first half of the trajectory (denoted in black) is used to predict the latter half of the trajectory. Since the two lines of WTTFNet with/ without WT overlapped in (c) and (d), both settings are merged to one line.

**E.2 Analysis of the role of weather and time-of-day**

In this section, further analysis on the role of weather and time-of-day in improving the prediction performance is studied. Following the significance $p$-value obtained for the McNemar's test in previous sub-section, which suggests that there is significant improvement in classification accuracy after adding WT information to the proposed approach. Moreover, 3008 pedestrians were found to have significant improvement after WT information were added. This motivates us to analyze the average displacement error (ADE) and final displacement error (FDE) of the 3008 pedestrians.

Figs. 5 and 6 compare the ADE and FDE of the proposed approach under two settings, respectively: with/without the incorporation of WT information. "No effect on Destination" means the predicted destination are same under the both settings, whereas "Influenced the Predicted Destination" means the predicted destination was altered after incorporating the WT information.

Fig. 5 shows the ADE of the proposed approach with/without WT information incorporated. From the figure, it can be shown that similar ADE was attained when the WT information has no effect on the predicted destination. On the other hand, if the predicted destination changed because of the varying weather (Influenced the predicted destination), the proposed WTTF with WT information incorporated will attain lower ADE (7.4083m) in compared to without WT information (7.83m). One-sided Mann–Whitney U test was used to test the significance in ADE reduction (7.83m to 7.4083m after adding WT information) and a $p$-value of $p = 0.0203 < 0.05$ was attained, suggesting the significance in performance improvement for these pedestrians considered.

Figure 6 shows the FDE under the two settings (with/without WT information) were compared for the proposed approach. Similar to the observation in the previous comparison, same FDE was attained when the WT information has no effect on the predicted destination (FDE=9.99m) and improved FDE (reduction from 14.11m to 13.04m) for the proposed WTTF approach when it changes the predicted destination after incorporating WT information. One-sided Mann–Whitney U test was used to test the significance in FDE reduction (14.11m to 13.04m after adding WT information) and a $p$-value of $p = 0.00533 < 0.05$ was attained, suggesting the significance in performance improvement for these pedestrians considered.

Overall, 5.47% (7.8m to 7.4m) and 7.58% (14.11m to 13.04m) improvement in ADE and FDE reduction were obtained for the 3008 pedestrians, and the reduction is found significant according to one-sided Mann–Whitney U tests. ($p$=0.0203 (<0.05) and $p$=0.00533 (<0.05) for ADE and FDE, respectively). For the remaining pedestrians, similar ADE and FDE performance was observed for pedestrians with no effect, because they have the same predicted destination under two settings (with/without WT information).

E.3 Qualitative Analysis

To illustrate the usefulness of adding WT information in the proposed WTTFNet, we consider four different cases, where Figs. 7(a) and (b) are extracted from the significant 3008 pedestrians and Figs 7(c) and (d) are extracted from the remaining pedestrians, whose destination was not affected by weather-time conditions.

Comparing between the proposed WTTFNet and other algorithms, the proposed WTTFNet (solid blue line with dots) generally aligns the best with the actual trajectory (solid black). In particular, the linear model, vanilla LSTM and PoPPL diverged inferiorly in Figs. 7(a) and 7(b).

To study the role of weather and time-of-day, we compare between the two different settings of the proposed WTTFNet: with/without WT information. From Figs 7(a) and 7(b), the WTTFnet with WT information (solid blue line with dots) aligns much better than the counterpart without WT information (solid red line with diamonds), which diverges in the middle of the path. For the remaining non-significant pedestrians, both settings nearly the same performance in Figs. 7(c) and 7(d) and hence only one of them are plot on the graphs.

Overall, the quantitative and qualitative analyses from Figs. 5 to 7 show that weather-time information helps to improve prediction performance significantly for the 3008 cases considered. Overall, the ratio 3008 out of 28536 pedestrians was also statistically significant according to the McNemar's test, suggesting that these 3008 pedestrians showing significantly improved performance out of 28536 cases were very unlikely a random event. This suggests the proposed approach may serve as an attractive approach for incorporating WT information to improve pedestrian trajectory prediction and it also serves as a systematic approach to test the significance of WT conditions.

## VI. CONCLUSIONS

A new deep WTTFNet has been presented. Experimental results using the Osaka ATC dataset [3] show that the proposed approach attained better performance than other state-of-the-art methods considered under varying weather-time conditions. A statistical test is also used to establish the significance of time-of-day and weather conditions. The proposed refinement framework can be adopted on other baseline models to improve these performance under varying weather-time conditions.

## REFERENCE

[1] J. Kantorovitch, J. Väre, V. Pehkonen, A. Laikari, and H. Seppälä, "An assistive household robot–doing more than just cleaning," *Journal of Assistive Technologies*, vol. 8, no. 2, pp. 64-76, 2014.

[2] S. Song et al., "Teleoperated robot sells toothbrush in a shopping mall: A field study," In *Extended Abstracts of the 2021 CHI Conference on Human Factors in Computing Systems*, pp. 1-6, May 2021.

[3] D. Helbing and P. Molnar, "Social force model for pedestrian dynamics", *Physical review E*, vol. 51, no. 5, 1995.

[4] M. Moussaïd, D. Helbing, and G. Theraulaz. "How simple rules determine pedestrian behavior and crowd disasters," *Proc. Nat. Acad. Sci.*, vol. 108, no. 17, pp. 6884-6688, 2011.

[5] F. A. Gers, J. Schmidhuber, and F. Cummins, "Learning to forget: Continual prediction with LSTM," *Neural computation*, vol. 12, no. 10, pp. 2451-2471, 2000.

[6] Y. Yao et al., "Bitrap: Bi-directional pedestrian trajectory prediction with multi-modal goal estimation," IEEE *Robotics and Automation Letters*, vol. 6, no. 2, pp. 1463-1470, 2021.

[7] (Liu et al. 2020) B. Liu et al., "Spatiotemporal Relationship Reasoning for Pedestrian Intent Prediction," *IEEE Trans. Robot. Autom. Lett*, vol. 5, no. 2, pp. 3485-3492, Apr. 2020.

[8] J. Wang, H. Sang, W. Chen, and Z. Zhao., "VOSTN: Variational One-shot Transformer Network for Pedestrian Trajectory Prediction," *Physica Scripta*, vol. 99, no. 2, 026002, 2024.

[9] A. Gupta, J. Johnson, L. Fei-Fei, S. Savarese and A. Alahi, "Social GAN: Socially acceptable trajectories with generative adversarial networks", *Proc. IEEE Conf. Comput. Vis. Pattern Recognit.*, pp. 2255-2264, Jun. 2018.

[10] Y. Yuan, X. Weng, Y. Ou, and K. M. Kitani, "Agentformer: Agent-aware transformers for socio-temporal multi-agent forecasting," *Proc. IEEE Conf. Comput. Vis. Pattern Recognit.*, pp. 9813-9823. June 2021.

[11] Z. Lv, X. Huang, and W. Cao, "An improved GAN with transformers for pedestrian trajectory prediction models," *Int. J. Intell. Syst.*, vol. 37, no. 8, pp. 4417-4436, 2022.

[12] Q. Du, X. Wang, S. Yin, L. Li and H. Ning, "Social Force Embedded Mixed Graph Convolutional Network for Multi-class Trajectory Prediction," to appear in *IEEE Trans. Intell. Vehicl.*, doi: 10.1109/TIV.2024.3352180.

[13] Z. He, T. Zhang, W. Wang, and J. Li, "A deep pedestrian trajectory generator for complex indoor environments," *Transactions in GIS*, vol. 28, no. 2, pp. 411-432, Apr. 2024.

[14] Y. Han, C. S. Tucker, T. W. Simpson, and E. Davidson, "A data mining trajectory clustering methodology for modeling indoor design space utilization," *Proc. International Design Engineering Technical Conferences and Computers and Information in Engineering Conference,* vol. 55898, V03BT03A017. American Society of Mechanical Engineers, 2013. https://doi.org/10.1115/DETC2013-12690

[15] M. D'Orazio, G. Bernardini, and Enrico Quagliarini, "How to restart? An agent-based simulation model towards the definition of strategies for COVID-19 second phase in public buildings," arXiv preprint arXiv:2004.12927[physics.soc-ph], 2020.

[16] Xue, Hao, Du Q. Huynh, and Mark Reynolds. "PoPPL: Pedestrian trajectory prediction by LSTM with automatic route class clustering." *IEEE Trans. Neural Netw. Learn. Syst.*, vol. 32, no. 1, pp. 77-90, 2020.

[17] A. Lui, Y. Chan, and K. Hung, "Functional Objects in Urban Walking Environments and Pedestrian Trajectory Modelling," *Sensors*, vol. 23, no. 10, 2023, pp. 4882.

[18] A. T. Steele, "Weather's effect on the sales of a department store," *Journal of Marketing*, vol. 15, no. 4, pp. 436-443, 1951.

[19] N. Rose, and L. Dolega, "It's the weather: Quantifying the impact of weather on retail sales," *Applied Spatial Analysis and Policy*, vol. 15, no. 1, pp. 189-214, 2022.

[20] E. Chung, O. Ohtani, H. Warita, M. Kuwahara and H. Morita, "Effect of rain on travel demand and traffic accidents," *Proc. 8th Int. IEEE Conf. Intell. Transp. Syst.*, pp. 13-16, 2005.

[21] H. Ren, Y. Song, S. Li, and Z. Dong, "Two-Step optimization of urban rail transit marshalling and real-time station control at a comprehensive transportation hub," *Urban Rail Transit*, vol. 7, pp. 257-268, 2021.

[22] Y. Shi, J. Xu, H. Zhang, L. Jia, and Y. Qin, "Walking model on passenger in merging passage of subway station considering overtaking behavior," *Physica A: Statistical Mechanics and its Applications*, vol. 585, pp. 126436, 2022.

[23] N. Tsiamitros et al.. "Pedestrian flow identification and occupancy prediction for indoor areas," *Sensors*, vol. 23, no. 9, pp. 4301, 2023.

[24] J. Peng, F. Peng, N. Yabuki, and T. Fukuda, "Factors in the development of urban underground space surrounding metro stations: A case study of Osaka, Japan," *Tunnelling and Underground Space Technology*, vol. 91, pp. 103009, 2019.

[25] A. Rasouli, I. Kotseruba and J. K. Tsotsos, "Are They Going to Cross? A Benchmark Dataset and Baseline for Pedestrian Crosswalk Behavior," *Proc. 2017 IEEE Proc. IEEE Int. Conf. Comput. Vis. Workshops* (ICCVW), Venice, Italy, 2017, pp. 206-213.

[26] J. Arevalo, T. Solorio, M. Montes-y-Gómez, and F. A. González. "Gated multimodal units for information fusion." arXiv preprint arXiv:1702.01992[cs], 2017.

[27] C. Li, M. Z. Zia, Q. Tran, X. Yu, G. D. Hager and M. Chandraker, "Deep supervision with intermediate concepts", *IEEE Trans. Pattern Anal. Mach. Intell.*, vol. 41, no. 8, pp. 1828-1843, Aug. 2019.

[28] T.-Y. Lin, P. Goyal, R. Girshick, K. He and P. Dollár, "Focal Loss for Dense Object Detection," *IEEE Trans. Pattern Anal. Mach. Intell*, vol. 42, no. 2, pp. 318-327, 1 Feb. 2020, doi: 10.1109/TPAMI.2018.2858826.

[29] D. Brščić, T. Kanda, T. Ikeda, and T. Miyashita, "Person tracking in large public spaces using 3-D range sensors," *IEEE Trans. Hum.-Mach. Syst.,* vol. 43, no. 6, pp. 522-534, 2013.

[30] Q. McNemar, "Note on the sampling error of the difference between correlated proportions or percentages," *Psychometrika*, vol. 12, no. 2, pp. 153–157, 1947.

[31] S. M. Ross, *Introduction to probability and statistics for engineers and scientists*. Academic press, 2020.

[32] H. B. Mann and D. R. Whitney, "On a Test of Whether one of Two Random Variables is Stochastically Larger than the Other," *Annals of Mathematical Statistics*, vol. 18, no. 1, pp. 50–60, 1947.

[33] Mehta, Vikas, and Jennifer K. Bosson. "Revisiting lively streets: Social interactions in public space." Journal of Planning Education and Research 41, no. 2 (2021): 160-172.

[34] X. Yang et al., "Stochastic user equilibrium path planning for crowd evacuation at subway station based on social force model," *Physica A: Statistical Mechanics and its Applications*, vol. 594, 127033, 2022.

[35] B. Sighencea, R. I. Stanciu, and C. D. Căleanu, "A review of deep learning-based methods for pedestrian trajectory prediction," *Sensors*, vol. 21, no. 22, pp. 7543, 2021.

[36] C. Zhang and C. Berger, "Pedestrian Behavior Prediction Using Deep Learning Methods for Urban Scenarios: A Review," *IEEE Trans. Intell. Transp. Syst.*, vol. 24, no. 10, pp. 10279 – 10301, 2023.

[37] D. Chowdhury, L. Santen, and A. Schadschneider, "Statistical physics of vehicular traffic and some related systems." *Physics Reports*, vol. 329, no. 4-6, pp. 199-329, 2000.

[38] A. Alahi, K. Goel, V. Ramanathan, A. Robicquet, L. Fei-Fei and S. Savarese, "Social LSTM: Human trajectory prediction in crowded spaces", *Proc. IEEE Conf. Comput. Vis. Pattern Recognit.*, pp. 961-971, 2016.

[39] A. Rasouli, I. Kotseruba, T. Kunic and J. Tsotsos, "PIE: A Large-Scale Dataset and Models for Pedestrian Intention Estimation and Trajectory Prediction," Proc. *2019 IEEE/CVF Int. Conf. Comput.Vis.* (ICCV), Seoul, Korea (South), 2019, pp. 6261-6270.

[40] F. Piccoli et al., "FuSSI-Net: Fusion of Spatio-temporal Skeletons for Intention Prediction Network," *Proc. 2020 54th Asilomar Conference on Signals, Systems, and Computers*, Pacific Grove, CA, USA, 2020, pp. 68-72.

[41] Y. Li, C. Zhang, J. Zhou and S. Zhou, "POI-GAN: A Pedestrian Trajectory Prediction Method for Service Scenarios," *IEEE Access*, vol. 12, pp. 53293-53305, 2024, doi: 10.1109/ACCESS.2024.3387698.

[42] A. Sadeghian et al., "Sophie: An attentive gan for predicting paths compliant to social and physical constraints," *Proc. IEEE Conf. Comput. Vis. Pattern Recognit.*, pp. 1349-1358. June 2019.

[43] A. Mohamed, K. Qian, M. Elhoseiny, and C. Claudel, "Social-STGCNN: A social spatio-temporal graph convolutional neural network for human trajectory prediction," *Proc. IEEE Conf. Comput. Vis. Pattern Recognit.*, pp. 14424–14432, June 2020.

[44] B. Yang et al., "Crossing or not? Context-based recognition of pedestrian crossing intention in the urban environment," *IEEE Trans. Intell. Transp. Syst.*, vol. 23, no. 6, pp. 5338-5349, 2021.

[45] Time and Date AS, "Historical Weather Data from Osaka International Airport, Japan," 2024. [Online]. Available: https://www.timeanddate.com. [Accessed: 2nd March, 2024].

[46] A. Paszke et al., "Pytorch: An imperative style, high-performance deep learning library," *Advances in neural information processing systems*, vol. 32 (2019).